# Empowerment Gain and Causal Model Construction: Children and adults are sensitive to controllability and variability in their causal interventions

Eunice Yiu[1], Kelsey Allen[2], Shiry Ginosar[3], and Alison Gopnik[1]

[1] Department of Psychology, University of California, Berkeley, CA 94720, USA
[2] Department of Computer Science, University of British Columbia, Vancouver, BC V6T 1Z4, Canada
[3] Toyota Technological Institute at Chicago, Chicago, IL 60637, USA

*Human knowledge and human power meet in one; for where the cause is not known the effect cannot be produced. Nature to be commanded must be obeyed; and that which in contemplation is the cause is in operation the rule.*
*—Francis Bacon*

Abstract

Learning about the causal structure of the world is a fundamental problem for human cognition. Causal models and especially causal learning have proved to be difficult for large pretrained models using standard techniques of deep learning. In contrast, cognitive scientists have applied advances in our formal understanding of causation in computer science, particularly within the Causal Bayes Net formalism, to understand human causal learning. In the very different tradition of reinforcement learning, researchers have described an intrinsic reward signal called "empowerment" which maximizes mutual information between actions and their outcomes. "Empowerment" may be an important bridge between classical Bayesian causal learning and reinforcement learning and may help to characterize causal learning in humans and enable it in machines. If an agent learns an accurate causal world model, they will necessarily increase their empowerment, and increasing empowerment will lead to a more accurate causal world model. Empowerment may also explain distinctive features of children's causal learning, as well as providing a more tractable computational account of how that learning is possible. In an empirical study, we systematically test how children and adults use cues to empowerment to infer causal relations, and design effective causal interventions.

Learning about the causal structure of the world is a fundamental problem for human cognition, and causal world models are central to intuitive theories. Such models allow wide generalization from limited data. Current large language models (LLMs), despite impressive progress in other areas, have very limited causal inference and learning capacities (Bender et al., 2021; Jin et al., 2024; Lewis & Mitchell, 2024; Kosoy et al., 2022). The inferential abilities they do display depend on the fact that they detect patterns in text and pictures that are generated by causal model-building humans (Yiu et al., 2024a, 2024b). Such a model might predict that the words "fire" and "smoke" are associated without any conception that fire causes smoke. In contrast, even very young children can and do spontaneously construct novel causal models of the world around them (e.g., Cook et al., 2011; Gopnik et al., 1999, 2004; Schulz & Bonawitz, 2007; for a review, see Goddu & Gopnik, 2024).

Where do these models come from in humans and how is this kind of learning possible? Over the last twenty-five years cognitive scientists have applied advances in our formal understanding of causation in philosophy and computer science to understand human causal learning. In particular, researchers have used the Causal Bayes net formalism which systematically relates directed acyclic causal graphs to patterns of conditional probability, interventions and counterfactuals (Pearl, 2000; Spirtes et al., 2000). It provides a natural way to describe both causal models and the patterns of data they generate.

The Bayes net formalism assumes an "interventionist" account of causation—roughly, variable X is causally related to variable Y if an intervention changing the value of X would lead to a change in the value of Y. This has also become a dominant account of causality in philosophy. Causation is distinctive, on this view, because it allows for successful action. Simply observing correlations or associations between events may allow you to predict one event from another. But an intervention to produce one event will only lead to another if there is a causal relation between them. The Bayes net formalism allows you to appropriately predict the consequences of interventions on variables as well as predicting the association between them.

This allows the reverse Bayesian inference, at least in principle. Given a pattern of outcomes of interventions and associations it should be possible to infer the causal model that was most likely to have generated that data. This approach to causal learning is part of a broader Bayesian probabilistic model approach to learning in developmental cognitive science (for a review, see Ullman & Tenenbaum, 2020). Empirically, causal learning in children is remarkably well captured by the Bayes net formalism (for reviews, see Gopnik & Wellman, 2012; Gopnik & Bonawitz, 2014; Goddu & Gopnik, 2024). Given a pattern of data, even preschool children infer the correct causal hypothesis.

From a computational perspective, however, inferring causal models from evidence, like many other kinds of Bayesian inference, proves to be intractable – the space of possible hypotheses is too large.

One approach might be to suggest that potential causal models are strongly constrained by prior innate "core knowledge" (e.g., Spelke, 2022). Although this might help address the search problem, it also undermines one of the major advantages of causal learning. Causal learning is so valuable precisely because it allows us to go beyond innate knowledge and learn about causal structure that may be counterintuitive—such as the workings of a TV remote or a smartphone. Causal learning even eventually allows us to revise our initial intuitive physics or psychology to produce scientific physics and psychology.

Another challenge for both Bayesian approaches and the classic "deep learning" algorithms that drive large pretrained models involves the role of active exploration and experimentation in causal learning. Children are not simply passive consumers of data. Instead, from very early on they actively seek out evidence that is relevant to the causal problems they are trying to solve. There has recently been extensive work showing just how motivated children and even infants are to experiment and explore, and how intelligently they do so (e.g., Du et al., 2023; Giron et al., 2023; Ruggeri et al., 2024, Schulz, 2012; Stahl & Feigenson, 2015). Similarly, in formal science, experimentation is the canonical way to discover causal structure. But, with a few exceptions, (e.g., Eberhardt & Scheines, 2007) there have not been computational accounts of how this kind of active experimentation takes place and how it allows causal inference.

In parallel, an apparently quite different kind of learning mechanism—reinforcement learning (RL)—has become increasingly influential in neuroscience and computer science, particularly when combined with modern "deep" machine learning (Dayan & Balleine, 2002; Sutton & Barto, 2018). For example, deep reinforcement learning has been the key to DeepMind's accomplishments in mastering Go and Chess (Silver et al., 2018).

As the philosopher James Woodward has pointed out, classical reinforcement learning might be considered a very specific and narrow form of causal learning (Woodward, 2007, 2021). In particular, it relates the specific actions an agent performs on the world—their interventions—to specific outcomes, in the form of external rewards. This basic structure is similar to the basic structure of causal relations on the interventionist account underlying the causal Bayes net formalism. Moreover, RL contrasts with other types of learning that are more purely associative, such as classical conditioning or neural network learning mechanisms. Associative learning captures correlations between variables and allows predictions about those variables, but doesn't allow a special role for interventions, and so is further removed from causal learning.

However, reinforcement learning typically applies only to the agent's own actions and to the rewards that follow those actions. More importantly, the basic motivational structure of reinforcement learning and causal learning are very different. Reinforcement learning is motivated by utilities – the attempt to maximize external rewards. Classically, causal learning is epistemically motivated – it involves approximating the true structure of the world. It is true that ultimately causal knowledge allows effective action and decision-making, and so increases utilities. But this is a long term and indirect effect. In the Bayesian framework, decision-making and utility calculations are layered on top of the fundamental epistemic project. The reverse is true in reinforcement learning—classic RL agents may try to learn about the environment, but they only do so in order to maximize further rewards later on.

Although RL methods can be very effective in relatively low-dimensional and fixed environments such as go or chess, they have much more difficulty in the high-dimensional, open-ended and non-stationary environments that are characteristic of human causal learning. In particular, RL systems face consistent difficulties in balancing exploitation—the fundamental utilitarian motivation of maximizing reward, and exploration—the more epistemic motivation of learning about the structure of the environment (e.g., Cohen et al., 2007; Sutton & Barto, 2018). In the long run, exploration will lead to more effective action, but it requires the agent to forego reward in the short run.

**Empowerment**

One approach to solving these problems has been to propose RL systems that seek internal epistemic rewards in addition to the classical external rewards. These systems implement the intrinsic epistemic motivation of classical causal learning in an RL framework. Several types of intrinsic rewards have been proposed in the literature. They include a variety of curiosity-based rewards, particularly measures of information gain, entropy and novelty (Oudeyer et al., 2007; Pathak et al., 2017; Schmidhuber, 2010). These rewards do seem to increase the exploratory efficacy of reinforcement learning systems. Moreover, there is evidence that children and even infants also seek out such intrinsic rewards, particularly information gain (Kidd & Hayden, 2015; Ruggeri et al. 2024; Schulz, 2012).

However, it is still difficult to sufficiently constrain these systems. For example, a classic failure mode for systems that seek out information gain is the "noisy TV" problem (Schmidhuber, 2010). Such systems will be captured by random noise like TV static. Seeking novelty and information gain by themselves doesn't seem to be adequate to understand the environment effectively in a way that supports action.

A different approach uses "empowerment" as an intrinsic epistemic reward (Klyubin et al., 2005, 2008; Salge et al., 2014). Interestingly, "empowerment" originally was formulated by Daniel Polani and colleagues in the evolutionary computation literature as part of an attempt to specify how intelligence might emerge even in very simple organisms. The crucial idea is that intelligence involves systematic relationships between "sensors"—systems for sensing and perceiving the environment—and "actuators" —systems for acting on the environment. A number of evolutionary theorists have pointed out that brains emerged in concert with the emergence of animals with coordinated perceptual and motor systems—such as eyes and claws—in the Cambrian explosion (Godfrey-Smith, 2020; Jablonka & Lamb, 2014). Empowerment described how sensors and actuators could be coordinated in an adaptive way.

In empowerment, an agent maximizes the mutual information between its actions and their outcomes, regardless of the reward value of those outcomes. In other words, the system is rewarded if variation in an action systematically leads to parallel variation in the outcome so that the value of the action variable predicts the value of the outcome variable. Simultaneously, the system is also rewarded for maximizing the variety of actions it takes, ensuring that it explores the widest range of possible actions and outcomes.Thus, seeking empowerment leads to both actions that allow more control of the environment and more variable actions.

Recently this idea has been applied to a variety of reinforcement learning problems (De Abril & Kanai, 2018; Du et al., 2020; Zhao et al., 2019). By endowing RL agents with empowerment as an intrinsic reward, those agents can explore and represent the environment more effectively. Rather than simply seeking out information, such a system will seek out exactly the relations in the world that allow the closest match between actions and outcomes—the most controllable relations—whether or not those outcomes are rewarding. This means that it will discover the relations that will be most likely to be useful for a wide range of future goal-directed actions. If I discover, for example, that moving a stick systematically changes the position of objects that it contacts, I can later pick up the stick to draw an out-of-reach object towards me. In fact, infants seem to learn to use sticks in just this way (Uzgiris & Hunt, 1975).

Although they come from different traditions, we argue that causal learning and empowerment gain are intimately related conceptually. If an agent learns an accurate causal model of the world, they will necessarily increase their empowerment, and, vice versa, increasing empowerment will lead to a more accurate (if implicit) causal world model.

This claim is rooted in the interventionist character of causation. One way we might think of causal learning is as an "inverse problem." An inverse problem involves inferring the structure of the external world from the data that world generates. A classic example is the way a visual system infers the structure of the three-dimensional world from retinal images. In inverse problems, we assume that there is, in fact, some single distinctive structure in the outside world that we are trying to reconstruct. A God's eye view of the universe would discover something corresponding to the 3D structure of objects. This structure is independent of the agents who are trying to understand it. Bayesian approaches in cognitive science formalize inverse problems in a probabilistic way.

However, the interventionist accounts that underlie causal Bayes nets (Pearl, 2000; Spirtes et al., 2000; Woodward, 2007) imply a different relation between agents and the world. TInstead, our notions of causation are rooted in the idea that causal relations are precisely those external relations that support effective interventions. In this tradition, we can define causal relations as those relations such that intervening to alter the value of a cause variable will lead to a corresponding change in the effect variable—in short, those relations where actions predictably produce outcomes. This is the same idea that is central to empowerment.

On the interventionist view, there may not be any single God's-eye-view, agent-independent causal structure analogous to, say, 3D spatial structure. The philosopher Peter Godfrey-Smith (2009) has argued that many ontologically disparate phenomena can all support causal interventions. There are many quite different relationships in the world—ranging from the intuitive physical relations that support "billiard ball" causation, to the belief-desire relationships of intuitive psychology, to the highly counterintuitive relations of scientific physics—that all happen to systematically support causal interventions. In all these cases, changing the value of one variable will systematically change the value of another, regardless of the mechanisms that allow this to happen. The relation between interventions and outcomes on this view is very close to the relation between actions and outcomes that constitutes "empowerment".

There are some differences between the actions in RL accounts of empowerment and the ideal interventions in Bayes net accounts. The ideal interventions that underpin causal inference are not identical to the intentional actions of actual agents. They are closely related, however, and in many circumstances agents' intentional actions will also serve as ideal interventions (see Woodward, 2020) and so pick out causal relationships in the world. Ideal interventions are exogenous: they are not themselves caused by the variables they intervene on; they directly influence the variables they intervene on and they do not independently influence downstream variables. In many cases, intentional actions on the world will also have these features and so justify causal inference.

Moreover, in more abstract and conceptual cases, the interventions may be theoretical rather than actual—to say that the moon causes the tides is to say that if we could alter the position of the moon, the tides would also alter, even if this intervention isn't actually possible. But if a relation is genuinely causal, this sort of intervention should at least be conceptually possible. Moreover, this distinguishes causal relationships from other relationships, such as spatial, geometric or logical relationships. And, in practice, the test for whether we have discovered causal relations is to perform experiments—to determine whether experimentally varying one variable will systematically predict the value of another, that is, precisely to look for high mutual information between actions and outcomes.

In sum, the basic structure of causal relations on the interventionist view is that we can intervene systematically on X to systematically change the value of Y, regardless of the specifics of X and Y. This is well captured by the notion of empowerment. So seeking empowerment gain will lead to the discovery of causal relations, and vice versa.

A further distinctive feature of both empowerment and causal learning, is that they can have both "mind to world" and "world to mind" directions of fit. It is possible to maximize an agent's empowerment by increasing its knowledge about how the world works (matching the mind to the world as in science). This is the classic picture of causal learning. However, it is equally possible to maximize empowerment by increasing an agent's skill and control over the world (matching the world to the mind as in engineering). This is more like the orientation of classic reinforcement learning. In engineering, we can actually introduce causal structure to the world rather than just detecting that structure. When we construct an artifact, like the gas pedal of a car, the relation between pressing the pedal and the subsequent movement will be highly empowering. Moreover, it seems right to say that pressing the pedal will cause the car to move, even if that causal relation didn't exist before we created it.

This difference in approach is also reflected in the formalisms that underpin classical causal learning and empowerment in RL. Causal discovery in the Bayes net tradition, as well as Bayesian inference more generally, assumes that the learner has an internal model and a stream of data, and the data reshapes the model. The model then supports interventions. But the interventions are causally isolated from the models themselves, they don't directly shape or otherwise influence the models, or vice versa, although they may serve as data for the causal learner. In fact, the assumption in Causal Bayes nets is that interventions are entirely exogenous, not caused by anything in the causal system, presumably including the causal model itself.

RL, in contrast, assumes that the learner is acting on the external environment, receiving feedback in the form of a reward signal, and modulating further actions as a result. In fact, in classic model-free RL the agent need not have any internal model of the environment at all. Even in model-based RL the model is imposed from outside rather than constructed through the RL process itself.

As a result, the typical picture of empowerment in reinforcement learning sees it as a feature of environmental states, the agent learns to move to states that will maximize empowerment—often described as moving to states that will allow a range of further actions. Brandel et al. for example, measured empowerment in human agents exploring a video game environment by determining whether people made objects that would be useful in creating further objects down-stream.

In fact, however, by far the most effective method for gaining more empowerment will be to construct a better internal causal model—as in the quote from Bacon at the start of this paper, science enables engineering, knowledge is power. But these internal models are not typically understood to be part of the state space in RL (although see Zhou et al., 2025 for recent work in this direction).

Conversely, a crucial part of causal model construction in both intuitive and scientific theorizing is to actively experiment on the environment, acting on the world in a way that is expressly designed to support causal model construction. But in the classic Bayes net formalism these interventions aren't themselves caused by features of the model.

The close conceptual link between empowerment and causality, then, is not entirely captured in the formalisms. In the causal Bayes net formalism, interventions sever the causal connection between the causal variable that is intervened on and earlier variables. But, in practice, and especially in experimentation, a causal model will actually determine which intervention the agent performs. Similarly, although reinforcement learning systems, including systems that maximize empowerment, may use externally imposed models, they do not typically have formal mechanisms that would allow them to construct such models from their actions. Bayesian causal learning depends on model-building rather than action, reinforcement learning depends on action rather than model-building. Human causal learning, even in infants, constantly goes back and forth between model-building and action. A formal bridge between Bayes net causal learning and empowerment gain in RL would be very desirable.

**Empowerment and the Psychology of Causal Learning**

Thinking about empowerment might also help us understand the psychology of causal learning. Recent papers in cognitive science suggests that adult exploration of an online environment can be well characterized by empowerment (Brändle et al., 2023, Melnikoff et al., 2022) and we have shown that this is also true for children (Du et al., 2023). Melnikoff et al. (2022) have also systematically shown that empowerment is related to the feeling of flow. But the empowerment approach more generally captures important features of early causal knowledge and learning and helps to explain a wide range of developmental findings.

Looking-time studies suggest that very young infants perceive some particular relations in intuitive physics that support causal inference, such as the relations in "billiard-ball" causality (Leslie, 1982). However, the development of causal concepts more broadly is initially closely linked to actions on the world and their outcomes. We have suggested (Goddu & Gopnik, 2024, following Woodward, 2000, and Grinfeld et al., 2020) that, both in phylogeny and ontogeny, causal understanding moves from a first-person perspective to a third-person perspective to an impersonal perspective. Reinforcement learning, which is found in almost all intelligent animals, represents a causal relation between the animal's own actions and their outcomes. Imitation learning, which is found in some non-human animals in limited ways, but is ubiquitous in human infants, represents a causal relation between another animal's actions and outcomes. The sort of impersonal causal understanding in science represents causal relations in the world independent of actual actions, though crucially supporting such actions in principle. Empowerment may be applied to all three types of relations.

There is empirical evidence that humans seek empowerment, and that this contributes to their causal learning. In the 70's, a series of papers suggested that even very young infants are indeed rewarded by empowerment. In classic studies of "conjugate reinforcement," Rovee-Collier (1979) tied a ribbon from a crib mobile to the infant's foot, so that kicking made the mobile move. Infants as young as 3 months old systematically acted to make the mobile move, varying their actions and observing the correlation between those actions and the behavior of the mobile.

In other studies infants would make a mobile move or activate a pattern of lights by turning their heads on a pressure sensitive pillow (Papousek & Papousek, 1975; Watson, 1972). Moreover, these actions could not simply be explained by classic reinforcement learning. Infants would learn to turn their heads or kick and would continue to do so, even though they no longer looked at the mobile or the lights, aside from a brief glance to check that their action was effective. Infants varied their actions and observed their results rather than simply converging on a single effective action. In addition, infants smiled and cooed when their actions consistently led to an effect, but not when that effect simply occurred independently of their actions (Watson, 1972). A more recent study with this methodology shows that infants consistently alter their actions on the mobile in a way that increases the contingency of their actions, again unlike classical reinforcement learning (Sloan et al., 2023). For example, infants systematically alternated between kicking and remaining still and observed the effects of both types of actions—a strategy that would increase empowerment, even though it would decrease immediate reward. In fact, Rovee-Collier described her results precisely as an empowerment reward: "*The control which the infants have gained over the consequences of their own actions seems to have become the reward, rather than the specific consequences per se.*" (Rovee-Collier, 1979).

In "conjugate reinforcement" infants are acting to maximize the empowerment of their own actions – their causal understanding has a first-person perspective. From early in infancy children also represent the goal-directed intentional actions of others and distinguish them from other kinds of events and movements (e.g., Woodward, 1998, 2009). Moreover, they map the goal-directed actions of others on to their own actions (Meltzoff, 2007). This is an important feature of human causal learning. It distinguishes it from other types of learning, such as classic reinforcement learning, which only concern an agent's own actions, and also distinguishes it from learning in other animals (Taylor et al., 2014). From early in life, then, children have the cognitive and conceptual structures in place to discover empowerment relations between actions and outcomes, both their own and those of others.

From at least 24 months and probably earlier, children make genuine causal inferences by observing the correlations between goal-directed actions—their own or others'—and outcomes. However, until around age 4, they do not make similar inferences from simple correlations between events (Bonawitz et al., 2010; Meltzoff et al., 2012; Waismeyer & Meltzoff, 2017). Suppose a 24-month-old sees a human hand repeatedly push a toy car against a block A, which causes a light to go on. Pushing the car against another block B does not have this effect. Now we ask the infant to make the light appear. Infants will reproduce the correct action on A in order to make the light go, but not the action on B. However, they will not do this if they simply see the car move on its own and cause the effect. This is true even though they will look towards the light in this condition, suggesting that they have learned the correlation between the motion of the car and the light (Meltzoff et al., 2012). In short, toddlers appear to detect empowerment relations between actions and outcomes and use those relations to infer causal relationships that determine their own future interventions. They do not do this based on correlations among events that do not involve actions and outcomes. Four-year-olds do infer new interventions from correlations alone, but this ability seems to depend on their earlier learning through goal-directed action.

Empowerment also naturally applies to children's early exploratory play (Chu & Schulz, 2020). Even infants characteristically play by varying their actions on an object and observing the results—hence the perennial popularity of toys like rattles and busy boxes that afford such empowering actions.

Empowerment based reinforcement learning, unlike Bayesian inference, also provides a natural way to characterize such experimental actions, they are precisely what you would expect from a system that was trying to act to maximize empowerment.

Thinking of causal learning in terms of empowerment may also help to resolve some of the search problems. Maximizing empowerment would not require the sort of search through a high-dimensional hypothesis space that is so challenging for Bayesian inference. Unfortunately, precisely calculating mutual information itself poses problems of tractability—but some approximation methods make such calculations more feasible (e.g., Zhao et al., 2020). Children might also begin by simply looking for correlations between actions, their own and others, and the outcomes that follow them, rather than fully calculating mutual information.

If children are maximizing empowerment, they would have a mechanism for independently discovering causal relations that are not specified innately, even without requiring the full apparatus of Bayesian causal inference. They might look for relations that have the feature of mutual information between their own actions and those of others, and outcomes, like the relations between sticks and distant toys. This might then allow them to build up a repertoire of basic causal arrows that can then be combined to build more complex models.

**Testing the Relations between Empowerment and Causality Empirically: The Star Machines**

We described a number of developmental studies suggesting that children seek empowerment gain in their everyday exploration. But could we test this idea more systematically? Empowerment involves two factors. Actions must lead to effects in the environment – they must enable control. But to obtain high empowerment, actions must also be variable. Again, this contrasts with simply seeking novelty or information in the environment. To maximize empowerment, and to explore the causal structure of the environment most effectively, you should seek a combination of control and variability.

In the following experiments, we systematically contrast empowerment gain, which involves both controllability and variability, with novelty, that is simple variability, on the one hand, or efficacy, that is simple control, on the other. A rotary dimmer dial is more empowering than a two-way light switch because it allows fine-tuned control over many more distinct brightness levels, rather than just two fixed states. However, if that same rotary dial functioned like a wheel of fortune, randomly determining an arbitrary brightness level with each turn, it would be less empowering than the basic light switch because despite providing a variety of outcomes, it lacks predictable control. Similarly, the rotary dial would yield more causal knowledge than the switch or the wheel of fortune. Will children and adults differentiate these cases and make appropriate causal inferences, interventions and generalizations as a result? Will they prefer events that afford empowerment over similar events with lower controllability or variability? We also ask whether children and adults have different preferences and make different interventions in some contexts and not others.

We conducted two studies examining whether children and adults recognize and prefer control, variability, or both, when making causal generalizations and interventions. Study 1 asked whether humans prefer pure controllability, pure variability, or their combination when reasoning about novel systems from observed evidence. After introducing three star machines that generated different kinds of star outputs (controllable but not variable, variable but not controllable, and controllable and variable), children were asked to select a machine to make various kinds of causal interventions and to keep in either goal-directed ("work") and exploratory ("play") contexts. Study 2 built on these findings by embedding two features of variation within a single machine—one feature varies predictably and the other randomly.

Study 1 Method

We introduced 80 five- to ten-year-old children ($\mu$ = 7.52 years, *SD* = 1.68 years) and 120 adults ($\mu$ = 27.57 years, *SD* = 4.30 years) to three machines, each designed to generate outcomes reflecting variability, controllability, or a combination of both. In a cover story participants were told that "the elf boss wants you to make stars with these machines." The machines were set up so that placing an object, initially a star, in one of several slots produced another different object (Figure 1).

### (a) Demonstration Phase

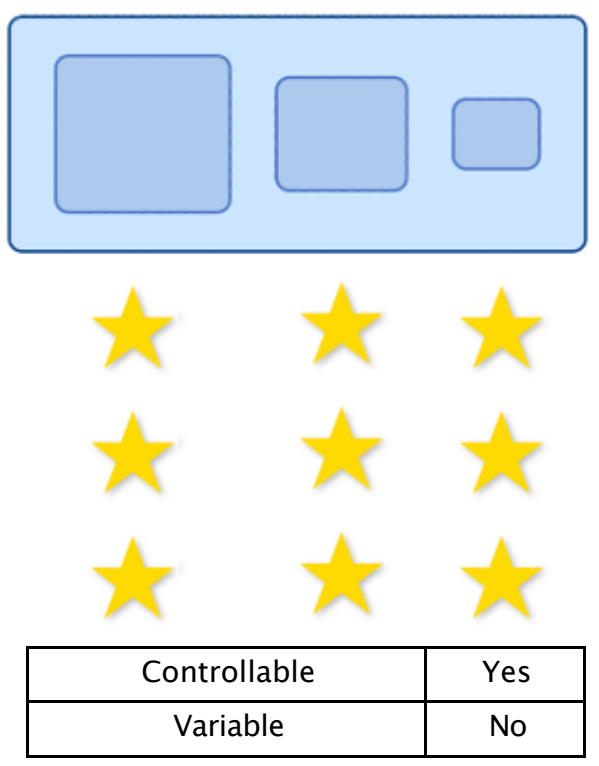

| Controllable | Yes |
| --- | --- |
| Variable | No |

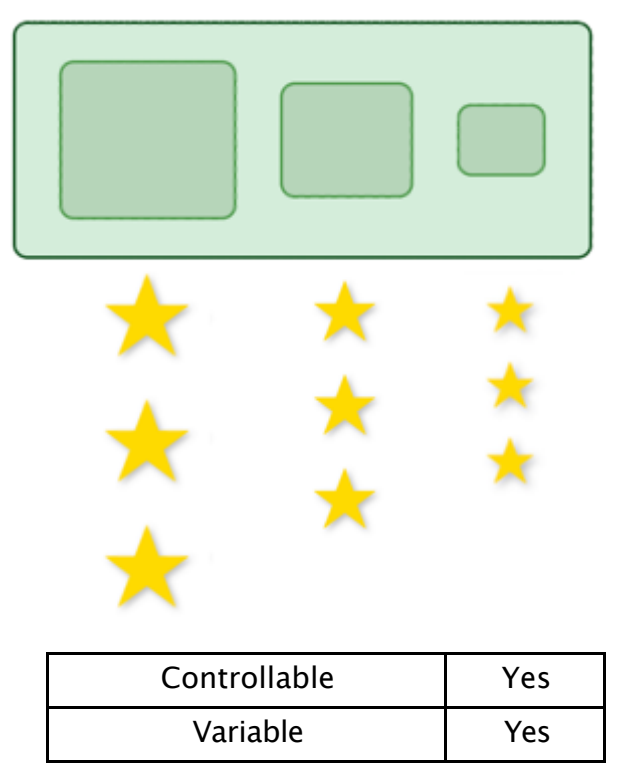

| Controllable | Yes |
| --- | --- |
| Variable | Yes |

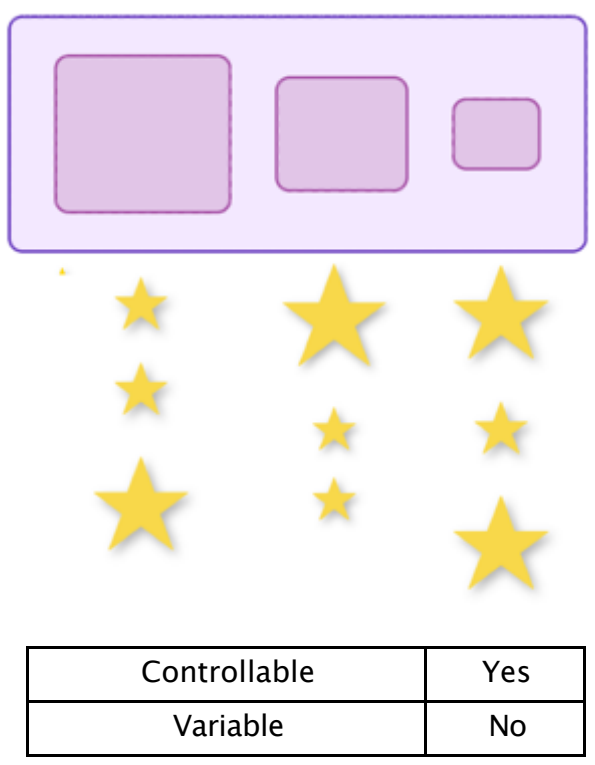

| Controllable | Yes |
| --- | --- |
| Variable | No |

Figure 1. Three machines characterized by the controllability and variability of their outputs. The purely controllable machine generates a single, deterministic output across all slots (left). The controllable and variable machine produces three distinct outputs, each reliably corresponding to slot size (middle). The purely variable machine generates three different outputs in a completely stochastic manner, with no predictable pattern (right). The color and order of the machines were fully randomized across participants.

Participants first watched how each machine produced outcomes. One machine (the purely controllable machine) always produced the same star size, regardless of the slot used. It either generated a large star or a medium star (60 adults and 46 children were randomly assigned to see it produce big stars, and the rest saw it produce medium stars). A second machine (the controllable and variable machine) showed a perfect correlation between slot size and output size: the big slot produced a big star, the medium slot produced a medium star (matching the input size), and the small slot produced a small star—an intuitive mapping where larger slots produced larger outputs. The third machine (the purely variable machine) generated star sizes randomly, with no relation between slot size and output size. The machines were otherwise identical, and their colors and positions were randomized across participants.

Participants were not told about the machines' underlying causal structure but had to infer it by observing the narrator drag stars into different slots. Every time a star was produced, a narrator commented on the change in size relative to the input: "Look it becomes smaller!", "Look it becomes bigger!", "Look it is the same!". Participants watched three outputs per slot, for a total of 27 outcomes across the three three-slot machines. Figure 1 illustrates an example demonstration. To eliminate memory demands, all the stars remained visible on the screen throughout the experiment.

### (b) Generalization Tests

After observing the demonstration, participants completed comprehension check questions identifying which star outputs came from which machine. They then moved on to the test phase, where they were presented with new causal generalization tasks. Critically, all the test tasks required them to generate new outcomes through their own actions.

We examined three levels of generalization: At the first level, participants had to extend the structure of the machine to a new output value. Participants were introduced to an "extra small" slot that was newly appended to each of the three machines and were asked to generate an "extra small" star that would be smaller than those they had previously observed. The correct choice was the extra small slot in the variable and controllable machine, rather than the slots in the other two machines (chance level = 1/3).

At the second level, participants had to apply the causal structure to a new object. They were given hats and were told to make three big, three medium and three small hats by using any slot on any of the three machines (twelve slots in total). Recall that some participants saw the purely controllable machine produce only large stars, whereas others saw it produce only medium stars. To make big hats, children in the first condition could use (1) any of the four slots on the purely controllable machine (which all produced large outputs) or (2) the large slot on the controllable and variable machine. In this condition, the combined chance level for selecting an appropriate slot was therefore 5/12. In the second condition, where the purely controllable machine produced only medium stars, children could rely only on the large slot in the controllable and variable machine, yielding a chance level of 1/12. The same logic applies to medium hats: when the purely controllable machine produced only medium stars, the chance level was 5/12 (four purely controllable slots plus the medium slot on the controllable and variable machine); when it produced only large stars, only the medium slot on the controllable and variable machine was appropriate, giving a chance level of 1/12. To make small hats, participants could only use the small or extra small slots on the controllable and variable machine, resulting in a chance level of 2/12.

At the third level, participants were asked to transfer the causal structure from object size to a new perceptual dimension – brightness. They were told that the machines could make light bulbs that were "bright," "sort of bright," "sort of dim," or "dim," and were then asked to produce one bright and one dim light bulb. This task was structurally identical to the previous hat task: making a bright light bulb followed the same logic as making a big hat, and making a dim light bulb mirrored making a small hat. The chance level for making a bright light bulb was either 5/12 or 1/12, depending on the participant's assigned condition (i.e., whether the purely controllable machine produced only large or only medium outputs), and 2/12 for making a dim light bulb.

(c) Machine Preference Test

The experiment concluded by asking participants which of the three machines they would keep if they were asked to "*work* to make new things" or if they could simply be "given more things to *play* with."

Study 1 Results

(a) Generalization Result

When asked to generalize to a new output value by making an extra small star, 46.25% of children and

80.83% of adults correctly selected the extra small slot on the variable and controllable machine (Figure 2(a)). This is significantly above chance (1/3) for both children ($p = .017$) and adults ($p < .001$) on a binomial test. Chi-square tests show the variable and controllable machine was significantly preferred by both children ($\chi2(2) = 13.9$, $p < .001$) and adults ($\chi2(2) = 166.4$, $p < .001$).

When asked to generalize to a new object by producing hats of different sizes, children selected the correct slots on the controllable machines 56.26% the time ($SE = 3.35\%$) and adults 86.85% the time ($SE = 1.90\%$) (Figure 2(b)). Across the nine trials, both groups showed a significant preference for the variable and controllable machine over the purely variable machine (children: $z = 6.24$, $p < .001$; adults: $z = 15.02$, $p < .001$) and over the purely controllable machine (children: $z = 3.15$, $p = .001$; adults: $z = 9.30$, $p < .001$). The preference for pure variability over pure controllability was significant only in children ($z = -3.09$, $p = .002$) but not in adults ($z = -5.72$, $p < .001$).

When asked to generalize to a new perceptual modality by creating light bulbs that alter in brightness instead of size, children and adults again succeeded in selecting the appropriate slots in the controllable machines. In the condition where the purely controllable machine produced only large objects, 71.74% of children and 76.67% of adults successfully created bright light bulbs significantly above the 5/12 chance level (binomial test, $p < .001$; Figure 2(c)(i)). In the condition where the purely controllable machine produced only medium-sized objects, participants had access to only one valid slot for producing a bright light bulb (the large slot in the variable and controllable machine), and yet 58.33% of children and 70% of adults still succeeded significantly above the 1/12 chance level (binomial test, $p < .001$; Figure 2(c)(ii)). Both children and adults (aggregated over the conditions where the purely controllable machine generates only medium outcomes or only large outcomes) showed a significant preference for the variable and controllable machine over the other two machines on a Chi-squared test (children: $\chi 2(2) = 10.2$, $p = 0.0061$; adults: $\chi 2(2) = 54.95$, $p < .001$) (Figure 2(c)(iii)). When making dim light bulbs, 48.10% children and 68.91% adults selected the correct slots, which again is significantly above the chance level of 2/12 (binomial test, $p < .001$). Children and adults significantly preferred the variable and controllable machine on a Chi-squared test (children: $\chi 2(2) = 17.4$, $p < .001$; adults: $\chi 2(2) = 108.12$, $p < .001$). There was no significant difference between the purely variable machine and the purely controllable machine.

Even though the purely variable machine could generate the full range of outcomes demanded by the three generalization tasks, children and adults rarely used this machine because these outcomes are uncontrollable and could not guarantee the desired outcome in a single intervention (20% and 5.04% respectively in making an extra small star; $\mu = 22.64\%$, $SE = 2.22\%$ and $\mu = 7.78\%$, $SE = 1.24\%$ in making nine different sized hats; $\mu = 22.15\%$, $SE = 3.31\%$ and $\mu = 15.06\%$, $SE = 2.32\%$ in making bright and dim light bulbs). This highlights humans' appreciation for controllability over variability in designing causal interventions for novel outcomes. It is interesting, however, that children showed more of a preference for pure variability than adults. This may simply reflect the greater noise in the children's responses, but it may also suggest that children are especially sensitive to information gain.

(a) Extra small output

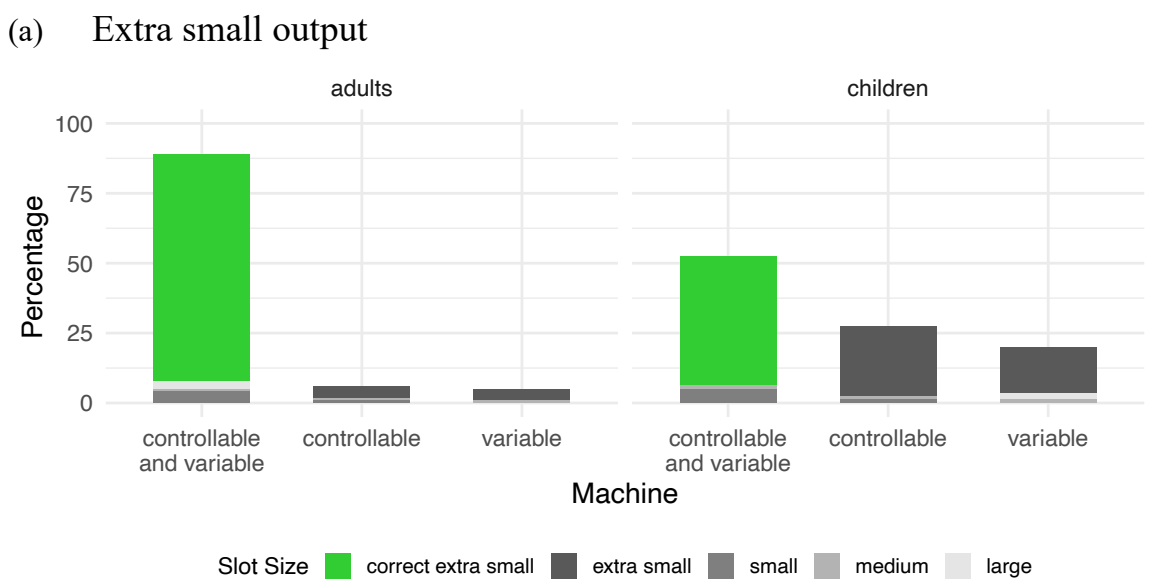

(b)(i) Big hats, controllable machine produces big objects

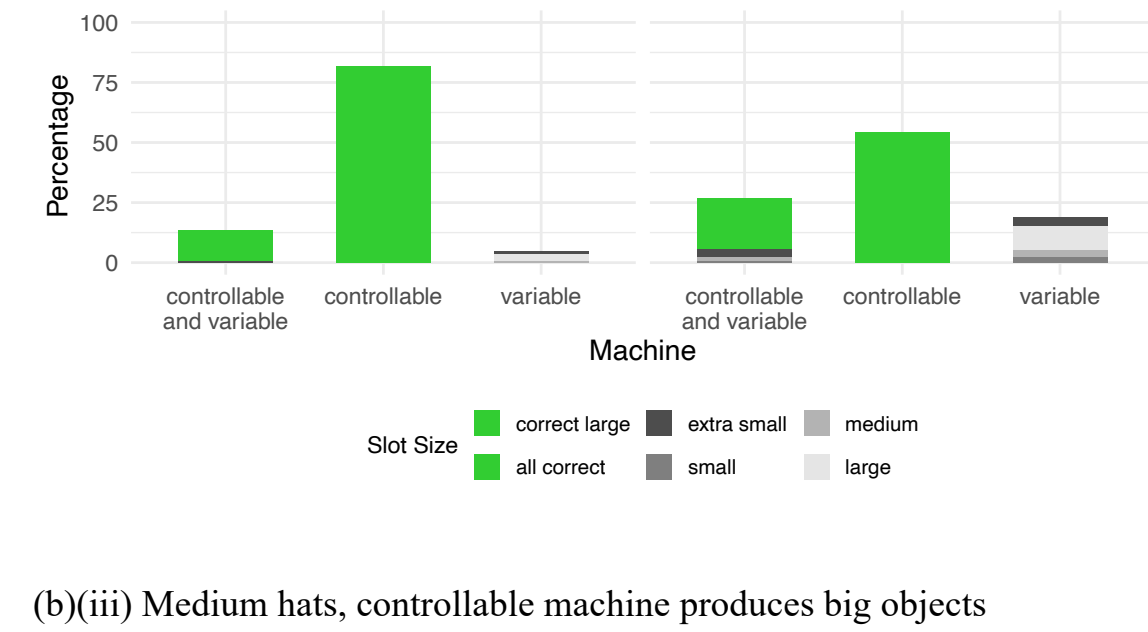

(b)(ii) Big hats, controllable machine produces medium-sized objects

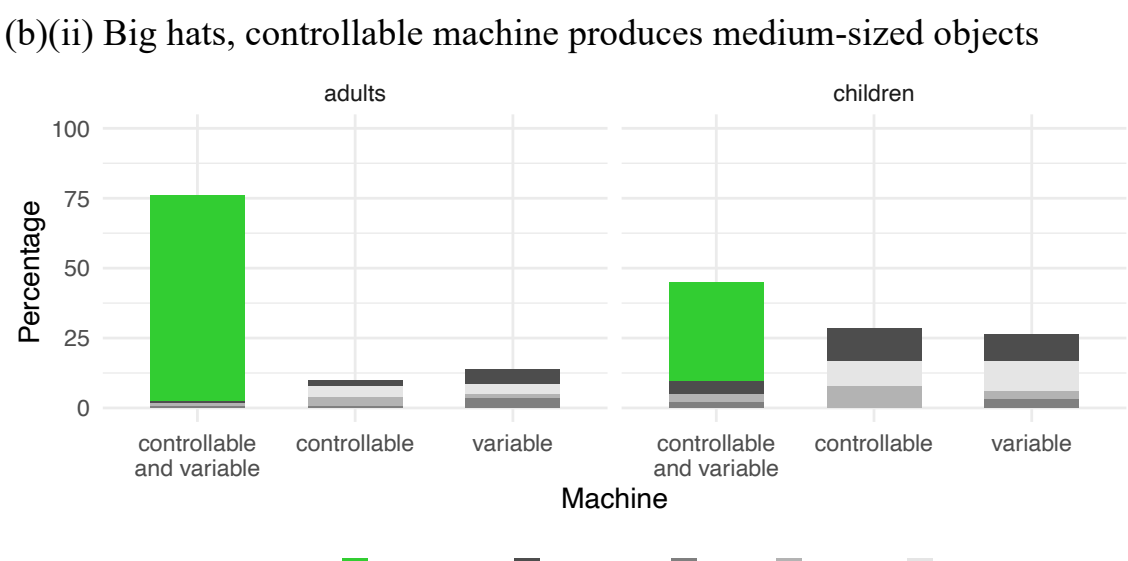

(b)(iii) Medium hats, controllable machine produces big objects

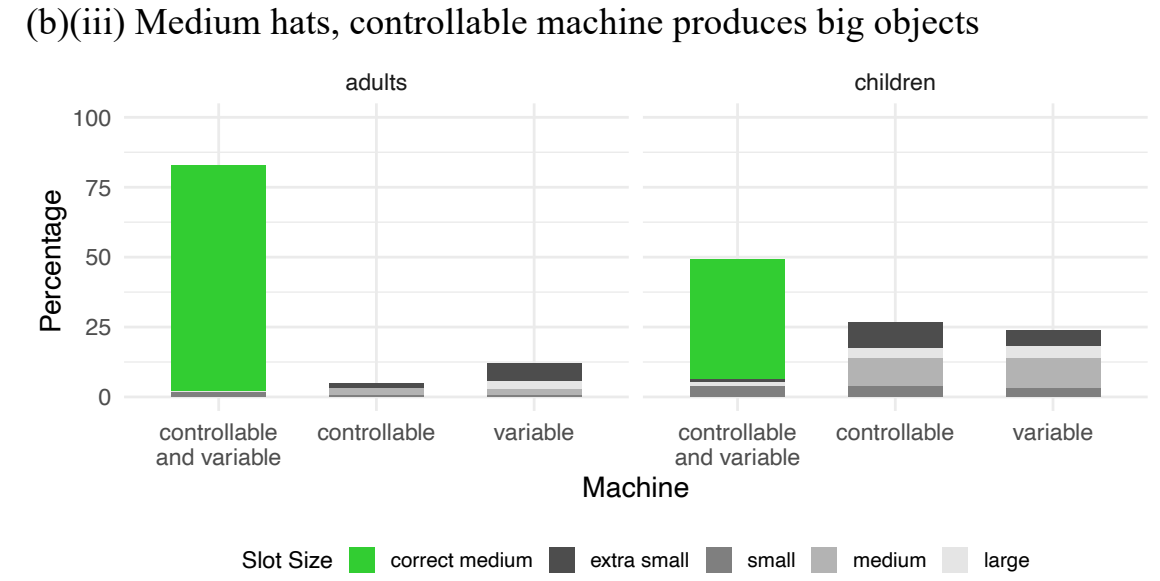

(b)(iv) Medium hats, controllable machine produces medium-sized objects

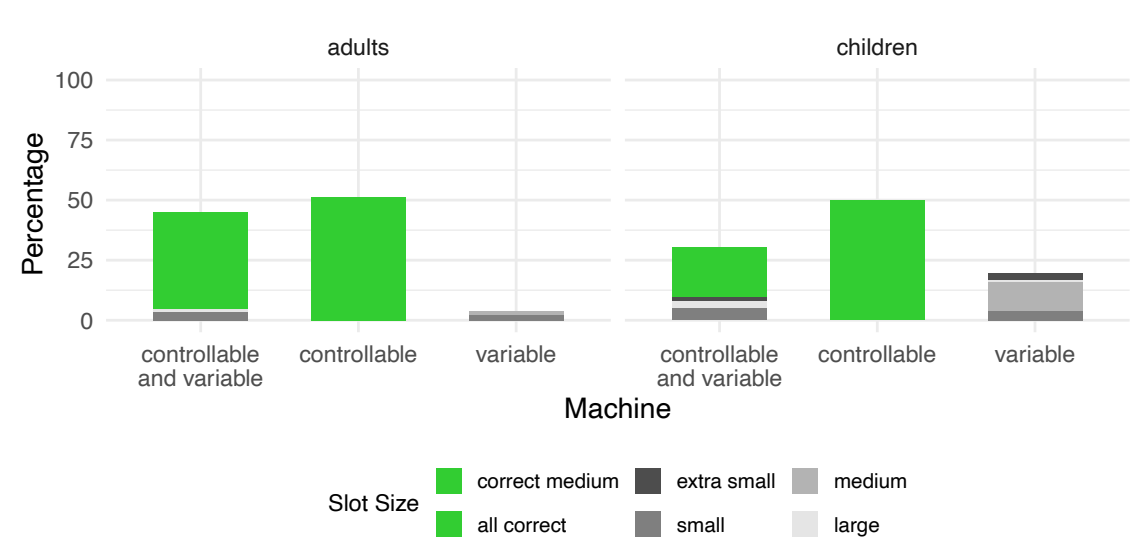

(b)(v) Small hats

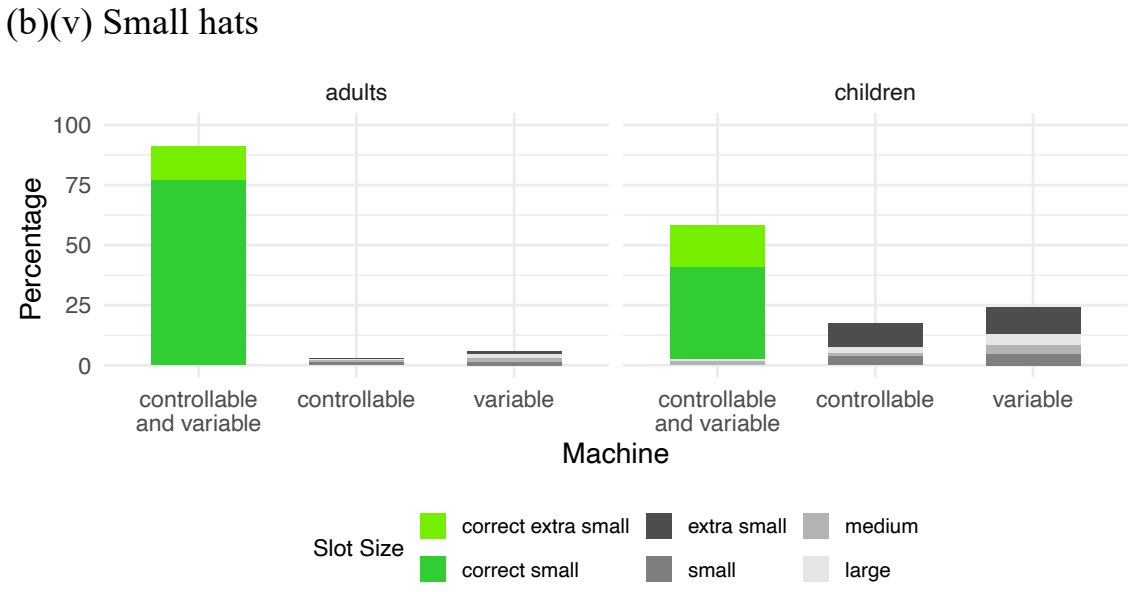

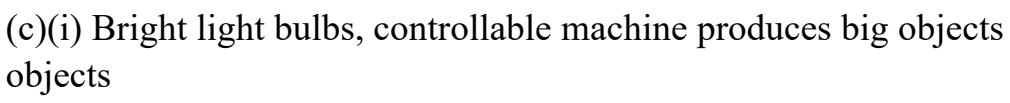

(c)(i) Bright light bulbs, controllable machine produces big objects objects

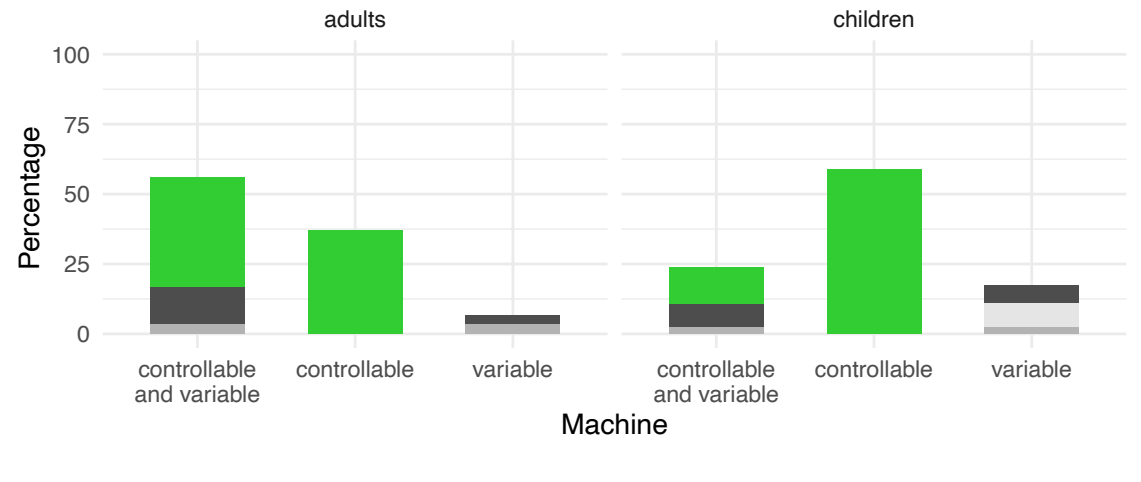

(c)(ii) Bright light bulbs, controllable machine produces medium-sized

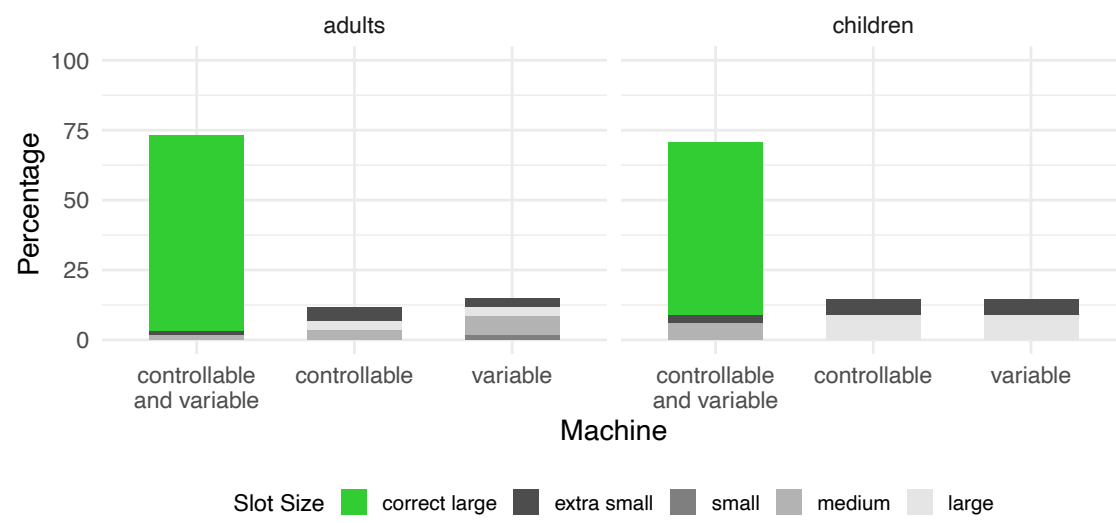

(c)(iii) Dim light bulbs

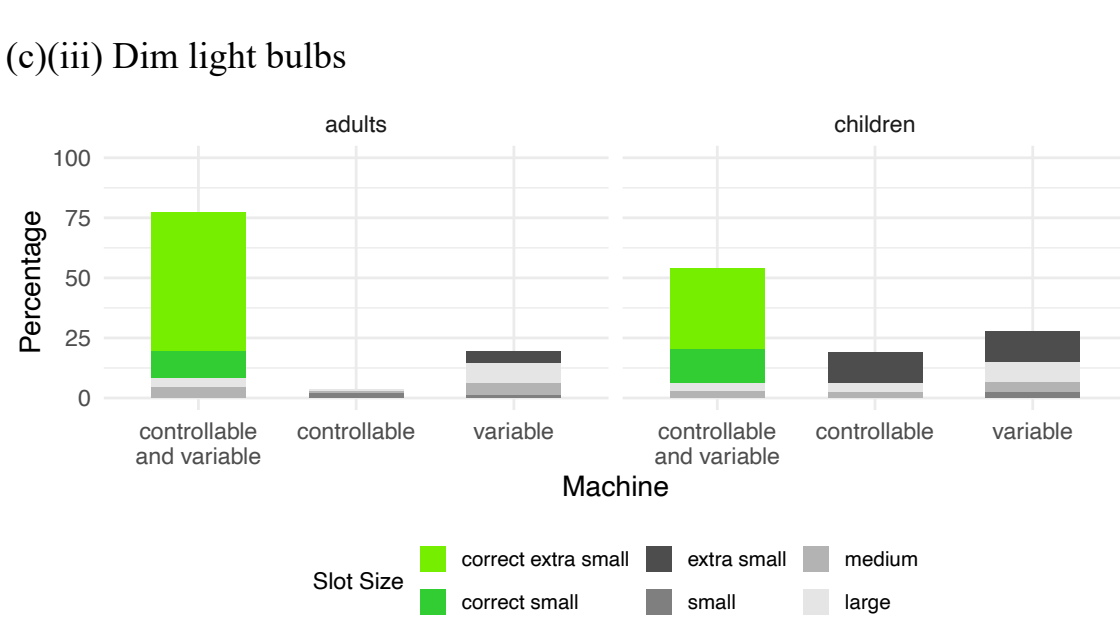

Figure 2. Distribution of machine and slot preference across the three generalization tasks: (a) to a new output value (extra small star), (b) to a new object kind (hats), (c) to a new perceptual modality (brightness).

## (b) Machine Preference Result

When asked to select a machine to keep for work, in particular, to make new kinds of things, the variable and controllable machine is most preferred (selected by 48.75% children and 75% adults), while the purely variable is least preferred (selected by 20% children and 66.7% adults) (Figure 3, left). A Chi-squared test revealed that children favored the variable and controllable machine (z = 2.93), preferred the purely controllable machine less (z = -.40, not significant), and avoided the purely variable machine (z = -2.53), $p$ = .009. Adults showed an even stronger preference for the controllable and variable machine (z = 9.68) and the other two machines were significantly under-selected (z = -3.49 for the purely controllable machine and -6.20 for the purely variable machine, $p < .001$ for both).

When asked to select a machine to keep for playing more, adults but not children continued to exhibit a strong preference for the controllable and variable machine (selected by 59.17% adults and 31.25% children) (Figure 3, right). Children's choices were evenly distributed on a Chi-squared test (variable and controllable: z = -.39, variable: z = .32, controllable: z = .079). By contrast, adults still strongly preferred the variable and controllable machine (z = 6.00) to the purely variable machine (z = -3.10, p < .001) and the purely controllable machine (z = -2.90, $p < .001$).

There was a significant shift in adults' machine preferences between the work and play contexts, $\chi2(2)$ = 10.94, $p$ = .012. Adults were significantly more likely to select the variable and controllable machine in the work context compared to the play context, $\chi2(1)$ = 5.68, $p$ = .017, while their preference for the purely variable machine increased in the play context, $\chi2(1)$ = 8.65, $p$ = .0033. There was no significant change in the selection of the controllable machine. Likewise, children demonstrated a significant shift in preference from controllability in the work context to variability in the play context, $\chi2(3)$ = 6.04, $p$ = .014. Like adults, children were significantly more likely to select the variable and controllable machine in the work context compared to the play context, $\chi2(1)$ = 6.04, $p$ = .014. At the same time their preference for the purely variable machine strengthened in the play context, $\chi2(1)$ = 5.04, $p$ = .025. In summary, children and adults show a stronger preference for the controllable and variable machine in the work context and shift to prefer the purely variable machine more in the play context.

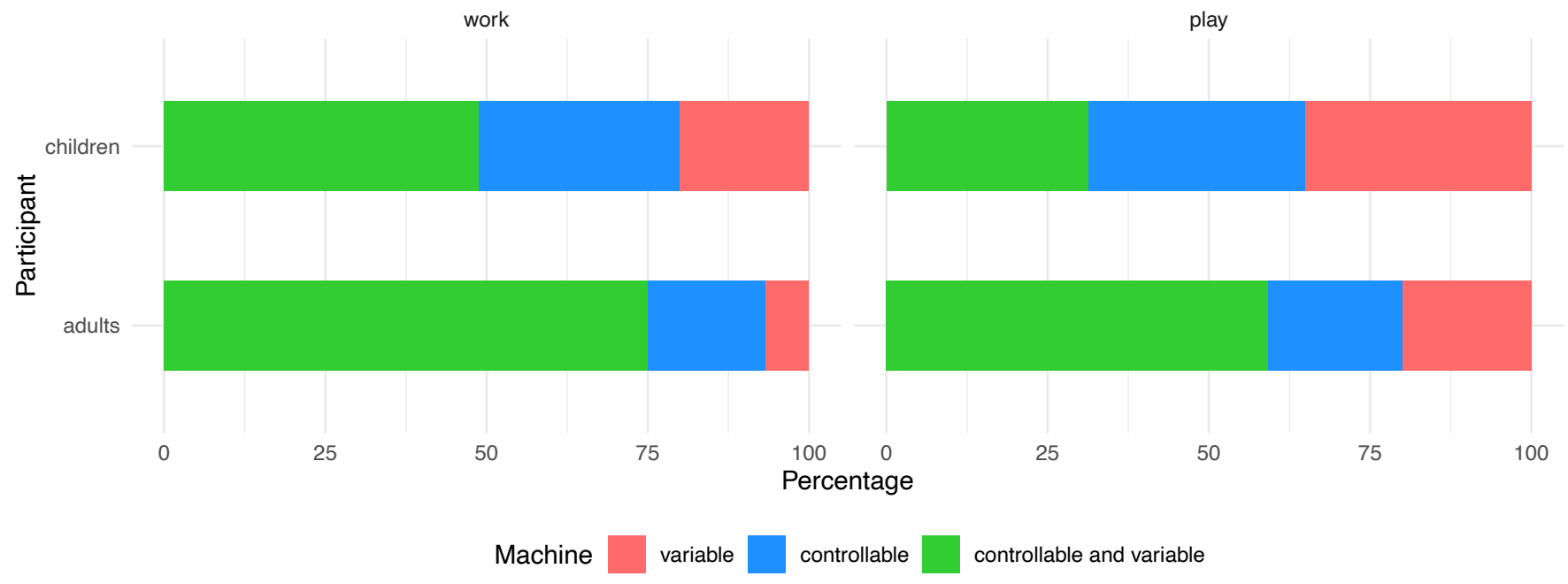

Figure 3. Proportion of machine selections by children and adults in the work and play contexts.

## Study 2 Method

In Study 1 children detected whether one machine was more controllable and variable than another. But different features of the same object could be also be empowering or irrelevant -the length of a stick will be correlated with spatial outcomes, its color will not. Can children detect and exploit the most empowering feature of an object—even when multiple features vary simultaneously? Study 2 was designed to replicate Study 1, but also to test this by contrasting two sources of variation (size and hue) within a single machine: one feature varies predictably while the other varies randomly. We ask whether children and adults selectively intervene on the more empowering feature to produce new outcomes.

We introduced 66 children (5-10 years old, *M* =7.88, *SE* = .21 years) and 60 adults (21-30 years old, *M* =26.70, *SE* = .35 years) to two machines, each designed so that one perceptual dimension followed a predictable pattern while the other varied randomly. Using the same cover story as in Study 1, participants were told that “the elf boss wants you to make stars with these machines.”

(a) Warm-up & Exploration Phase

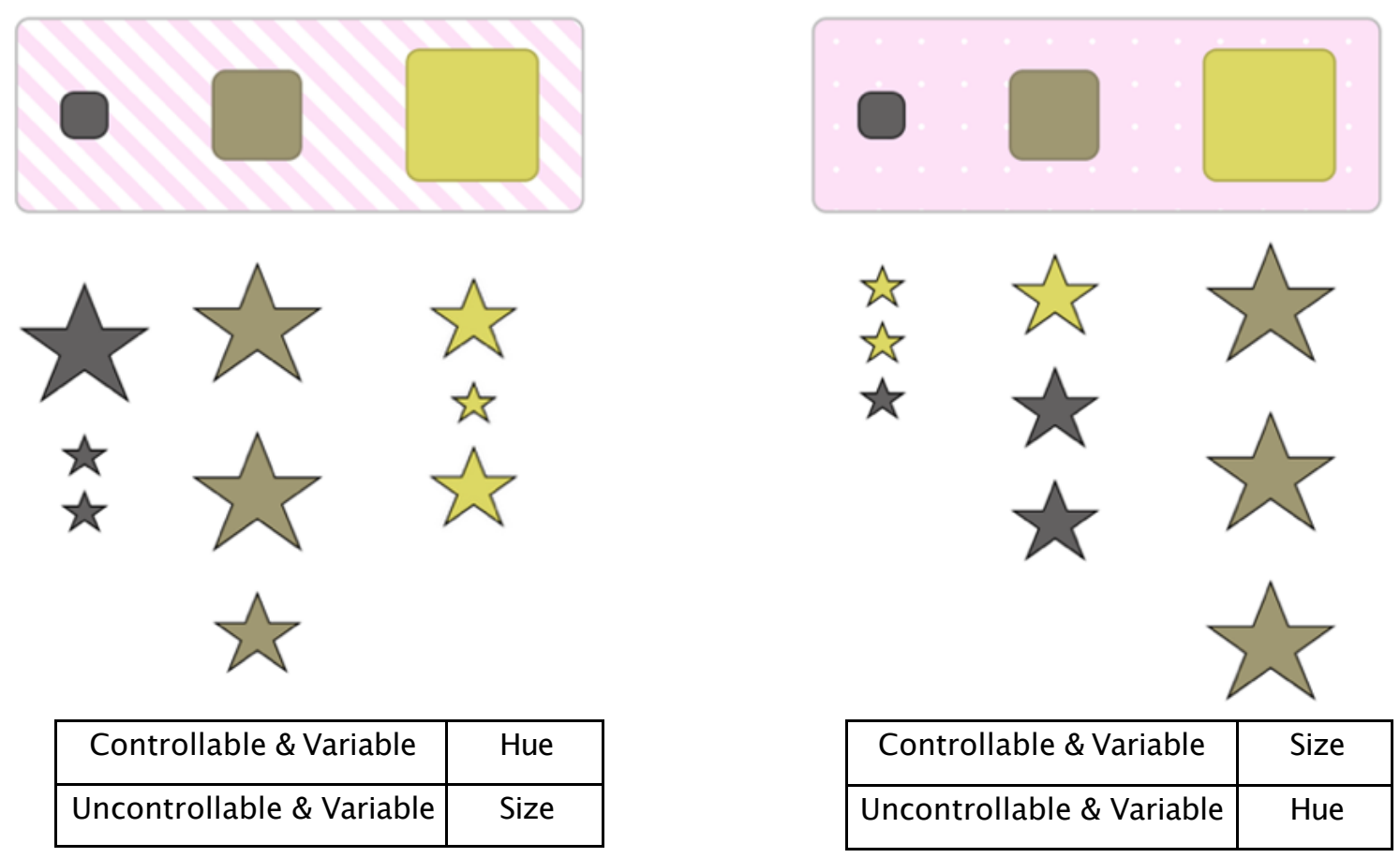

Figure 4. Two machines combining controllable and uncontrollable (random) variation across three slots. Hue Machine (left): Hue is controllable; size is not. Size Machine (right): Size is controllable; hue is not.

To ensure that participants understood the relevant perceptual features, participants were first asked to point to the “darkest,” “brightest,” “biggest,” “smallest” and “in-between” stars before observing the machines. Next, as in Study 1, participants played the role of an elf operating a star factory with two machines (Figure 4). In one machine, *hue (brightness)* varied systematically while *size* varied randomly (Figure 4, left); in the other, this mapping was reversed (Figure 4, right). Crucially, whenever a dimension was randomized, its three levels were evenly represented—participants saw three dark, three medium, and three bright stars (or three large, three medium, and three small stars) in random order. Participants received 18 stars and, in sets of three, chose one of six slots in which to drop the stars—using all 18 stars and placing no more than three in any single slot. This procedure promoted active engagement and attention to each output transformation, ensured that all participants observed the same amount of evidence (18 outputs), and allowed direct comparison of outputs produced by the same slot.

(b) Generalization Tests

After exploring the machines, participants were randomly assigned to either a size or hue condition and then completed two tasks: extending the mapping to a new output value and applying the mapping to a new object. Participants were introduced to a new "extra large" and "extra bright" slot that was appended to each of the two machines and were then asked to make an extra large star, or an extra bright star.

In the new object condition, participants were presented with two machines with three slots each. They had to create a big and a small hat respectively in the size condition and a bright and a dark light bulb respectively in the hue (brightness) condition.

(c) Machine Preference Test

Finally, we asked participants which of the two machines they would keep if they were asked to "make bigger things" versus "make brighter things."

Study 2 Results

(a) Generalization Result

When asked to make an "extra large" star, 45.50% of children and 63.33% of adults correctly selected the extra large slot on the size machine (see Figure 5(a)(i)). This is significantly above chance (1/8) for children and adults ($p < .001$) on a binomial test. When asked to make an "extra bright" star,42.40% of children and 70% of adults correctly selected the extra bright slot on the hue machine (see Figure 5(b)(i)), which is significantly above chance (1/8) for both groups.

59.10% children and 71.67% adults correctly selected the large slot on the size machine to make large hats (see Figure 5(a)(ii)), while 53.03% children and 65% adults correctly chose the small slot on the size machine to make small hats (see Figure 5(a)(iii)). When asked to make light bulbs of different brightness levels, 54.55% children and 76.67% adults correctly selected the bright slot on the hue machine to make bright light bulbs (see Figure 5(b)(ii), whereas 54.55% children and 58.33% adults correctly chose the dark slot on the hue machine to make dark light bulbs (see Figure 5(b)(iii)). All slot selections were significantly above chance (1/6) ($p < .001$), and the relevant feature-controllable machine is preferred significantly ($p < .05$) by both groups.

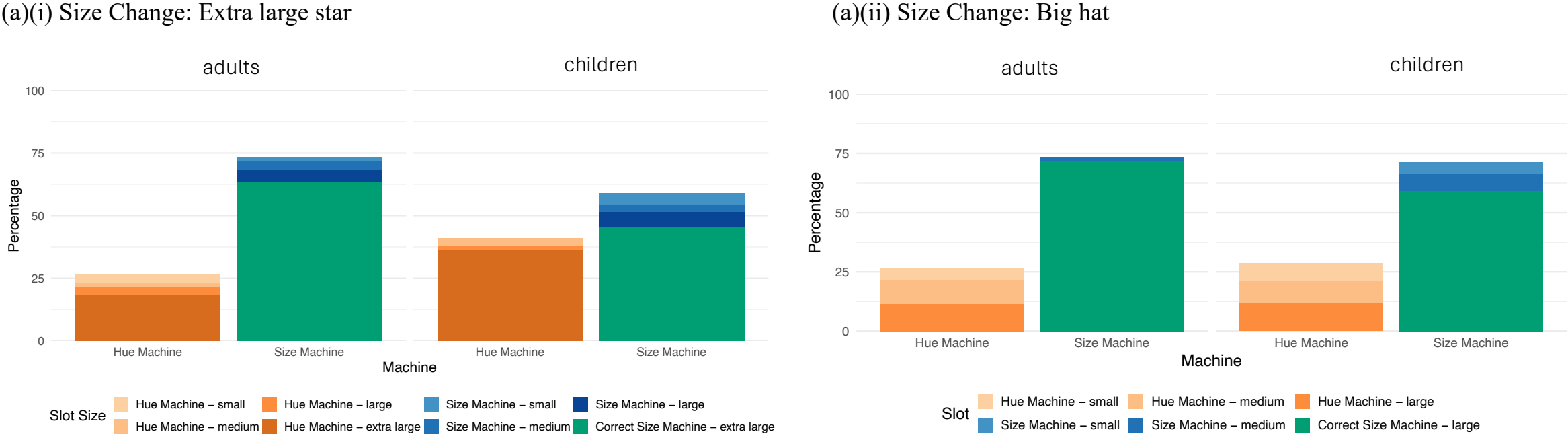

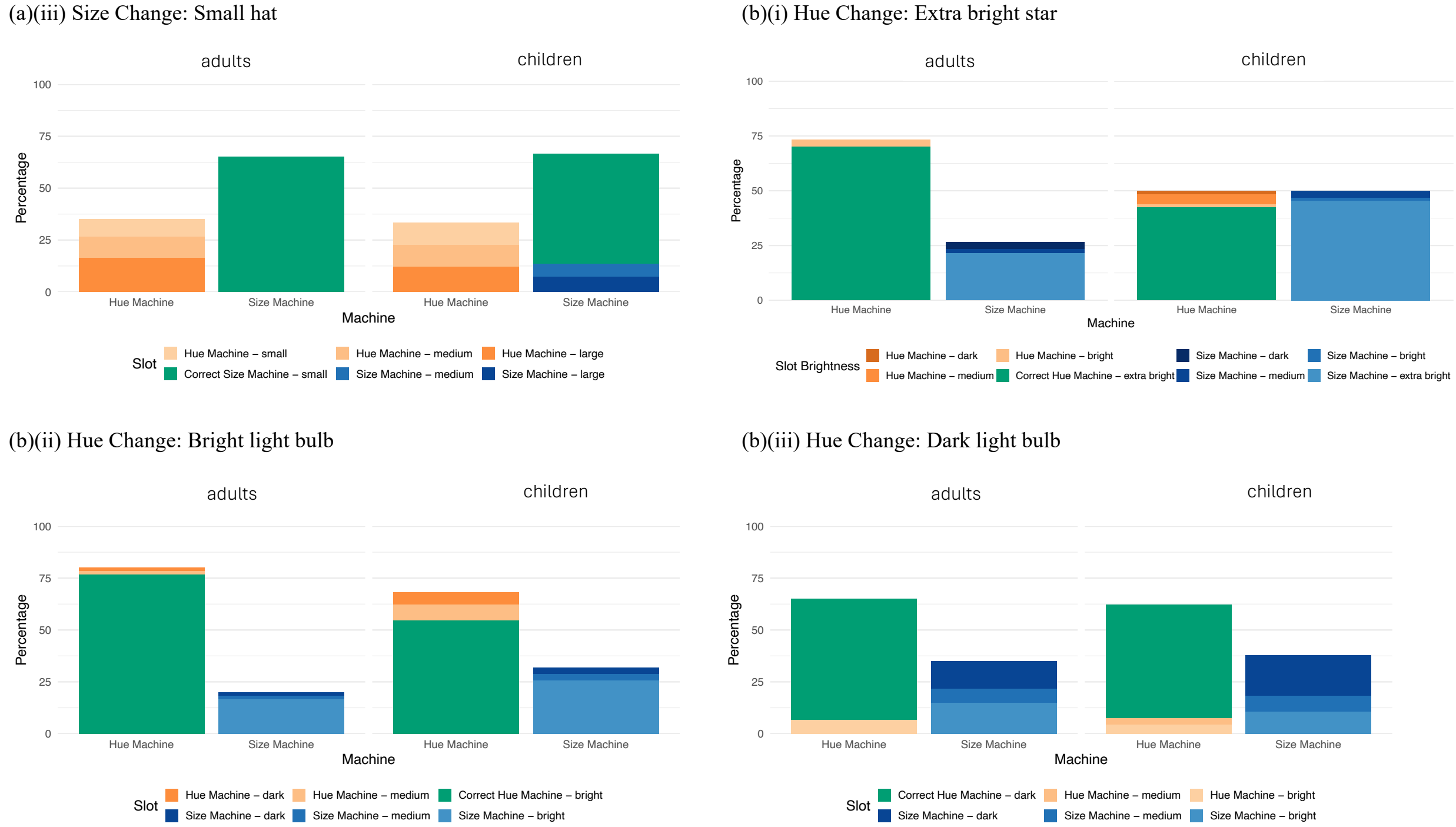

Figure 5: Distribution of machine and slot preference across the generalization tasks in the (a) size condition: (i) to a new output size value (extra large star) and (ii)-(iii) to a new object (hats); and in the (b) hue condition: (i) to a new output size value (extra bright star) and (ii)-(iii) to a new object (light bulbs).

## (b) Machine Preference Result

When asked to select a machine to "make things bigger," 72.73% children and 80% adults opted for the size machine (see Figure 6, left). When asked to select a machine to "make things brighter," 71.21% children and 88.33% adults selected the hue machine (see Figure 6, right). In other words, both groups showed a statistically significant explicit preference for the feature-controllable machine to make thing that that vary in the queried feature ($p < .01$ on binomial test).

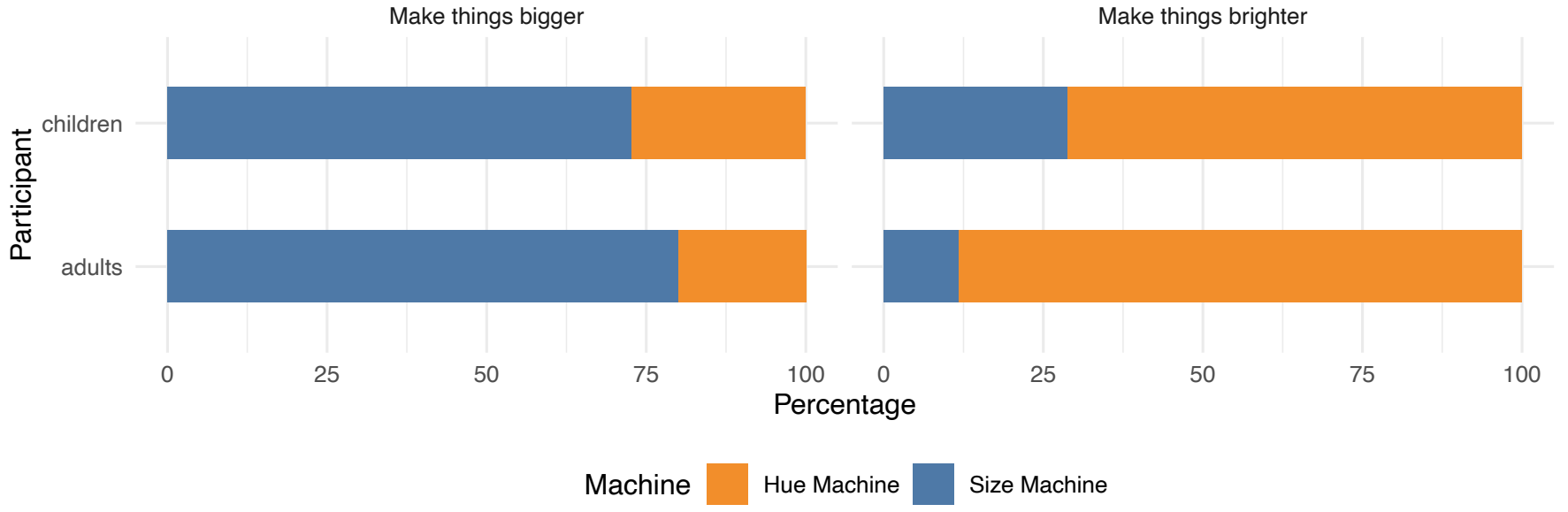

Figure 6: Proportion of machine selections by children and adults for making new things different in hue and in size respectively.

## General Discussion

Across two studies, adults and child learners sought out causal systems in which their actions produced the most variable and controllable effects. Study 1 established the basic distinction between controllability (actions reliably change outcomes) and variability (outcomes differ). When these two properties were cleanly separated across machines, both children and adults preferred the machine that offered *controllable variability* for goal-directed causal intervention. Both children and adults, however, shifted more towards the random, purely variable machine when they decided what to play with.

Study 2 extended this logic to a more complex scenario in which multiple features varied simultaneously, but only one had a systematic causal outcome. Here, participants needed not only to detect reliable structure but to *ignore* another salient, but uncontrollable, source of variability. Children and adults attended to and selected the dimension with high mutual information between actions and outcomes, generating new outcomes along that dimension and explicitly preferring the machine that made the feature more controllable.

Our behavioral results show that even children search for, prefer, and generalize from mappings that enhance their potential to control outcomes, effectively using empowerment gain as a guide for constructing their causal models. When participants infer reliable causal structure (mind-to-world), they increase their empowerment. When they act to exploit controllable dimensions (world-to-mind), they reinforce and refine their causal model. The two processes bootstrap one another: building a causal model grants greater empowerment, and acting to maximize empowerment reveals causal structure.

## Conclusion

In sum, we argue here that recent work on "empowerment" may help bridge Bayesian and RL approaches to learning and provide both empirical and theoretical insight into the crucial problem of learning the causal structure of the world. We also present a first set of experiments exploring this idea empirically.

Acknowledgements and Support: This research was supported by grants from the Google-BAIR Commons, DARPA Machine Common Sense Program, the ONR MURI program, the Templeton Foundation, the Templeton World Charity Foundation and CIFAR. Yu Qing Du, Li Dayan, Eliza Kosoy, Caren Walker, Daphna Buchsbaum, Kasra Jalaldoust, and Ella Qiawen Liu were central to these ideas and the empirical research and discussions with Nihat Ay, Daniel Polani, Pieter Abeel, Charles Kemp, Andrew Perfors, Lerrel Pinto, Yann LeCun, and Tom Everitt - along with members of BAIR, the Ranch Metaphysics Workshop, and the Nature of Intelligence Working Group at Santa Fe Institute - all contributed greatly to clarifying these ideas. Finally, we are also grateful to the participants and parents, as well as the Berkeley undergraduate students who assisted, Anisa Noor Majhi, Janna Umagat, Kaydee Manikhong, Ray Huang, Shivalika Jhabua and Srinitya Sriram.